\documentclass[letterpaper]{article} 
\usepackage{aaai24}  
\usepackage{times}  
\usepackage{helvet}  
\usepackage{courier}  
\usepackage[hyphens]{url}  
\usepackage{graphicx} 
\usepackage{natbib}  
\usepackage{caption} 
\usepackage{algorithmic}
\usepackage[ruled,vlined]{algorithm2e}
\usepackage{float}

\usepackage{amsmath,amsfonts,bm}

\def\eqref#1{equation~\ref{#1}}

\def\1{\bm{1}}

\def\mM{{\bm{M}}}

\def\mW{{\bm{W}}}
\def\mX{{\bm{X}}}
\def\mY{{\bm{Y}}}

\DeclareMathAlphabet{\mathsfit}{\encodingdefault}{\sfdefault}{m}{sl}
\SetMathAlphabet{\mathsfit}{bold}{\encodingdefault}{\sfdefault}{bx}{n}

\newcommand{\KL}{D_{\mathrm{KL}}}

\DeclareMathOperator*{\argmin}{arg\,min}

\usepackage{newfloat}
\usepackage{listings}
\DeclareCaptionStyle{ruled}{labelfont=normalfont,labelsep=colon,strut=off} 
\floatstyle{ruled}
\newfloat{listing}{tb}{lst}{}
\floatname{listing}{Listing}
\usepackage{multirow}
\usepackage{bbding}
\usepackage[table,xcdraw]{xcolor}

\title{Fast and Controllable Post-training Sparsity: \\ Learning Optimal Sparsity Allocation with Global Constraint in Minutes}
\author{
    Ruihao Gong\textsuperscript{\rm 1,2},
    Yang Yong\textsuperscript{\rm 2},
    Zining Wang\textsuperscript{\rm 1},
    Jinyang Guo\textsuperscript{\rm 3,1},
    Xiuying Wei\textsuperscript{\rm 2},
    Yuqing Ma\textsuperscript{\rm 3,1},
    Xianglong Liu\textsuperscript{\rm 1}\thanks{Corresponding author.}
}
\affiliations{
    \textsuperscript{\rm 1}State Key Laboratory of Complex \& Critical Software Environment, Beihang University \\
    \textsuperscript{\rm 2}SenseTime Research\\
    \textsuperscript{\rm 3}Institute of Artificial Intelligence, Beihang University\\
    \{gongruihao, 19373122, jinyangguo, mayuqing, xlliu\}@buaa.edu.cn, \{yongyang, weixiuying\}@sensetime.com

}

\usepackage{bibentry}

\begin{document}

\maketitle

\begin{abstract}
Neural network sparsity has attracted many research interests due to its similarity to biological schemes and high energy efficiency. However, existing methods depend on long-time training or fine-tuning, which prevents large-scale applications. Recently, some works focusing on post-training sparsity (PTS) have emerged. They get rid of the high training cost but usually suffer from distinct accuracy degradation due to neglect of the reasonable sparsity rate at each layer. 
Previous methods for finding sparsity rates mainly focus on the training-aware scenario, which usually fails to converge stably under the PTS setting with limited data and much less training cost. 
In this paper, we propose a fast and controllable post-training sparsity (FCPTS) framework. By incorporating a differentiable bridge function and a controllable optimization objective, our method allows for rapid and accurate sparsity allocation learning in minutes, with the added assurance of convergence to a predetermined global sparsity rate. 
Equipped with these techniques, we can surpass the state-of-the-art methods by a large margin, e.g., over 30\% improvement for ResNet-50 on ImageNet under the sparsity rate of 80\%. Our plug-and-play code and supplementary materials are open-sourced at \url{https://github.com/ModelTC/FCPTS}.
\end{abstract}

\section{Introduction}
Deep neural networks (DNNs) have achieved remarkable success in a variety of fields, including computer vision, natural language processing, and information retrieval. However, when deploying DNNs on resource-limited edge devices, reducing the memory footprint of neural networks and improving energy efficiency become crucial problems. Therefore, various compression techniques are proposed to make the models efficient~\cite{tan2019mnasnet,2021-neuips-mqbench,2022-neurips-outlier_suppression,hinton2015distilling,jacob2018quantization,2019-iccv-dsq,li2021brecq,2020-cvpr-int8training}. Among these compression techniques, model sparsity, which prunes the unimportant weights to zero, relates most with the biological brains. Several studies~\cite{hoefler_sparsity_2021, jacob2018quantization, gopalakrishnan2018combating,liu2020autocompress,liu2021lottery,yuan2021mest,hu2023channel} demonstrate that sparsity can contribute to a more robust and generalized model. Additionally, sparsified weights can be stored in special formats that consume less memory. Due to the advantage of low inference cost, low memory requirement, and high generalization ability, model sparsity has attracted much research interest in the community.

\begin{figure}[t!]
	\begin{center}
		\includegraphics[width=0.87\linewidth]{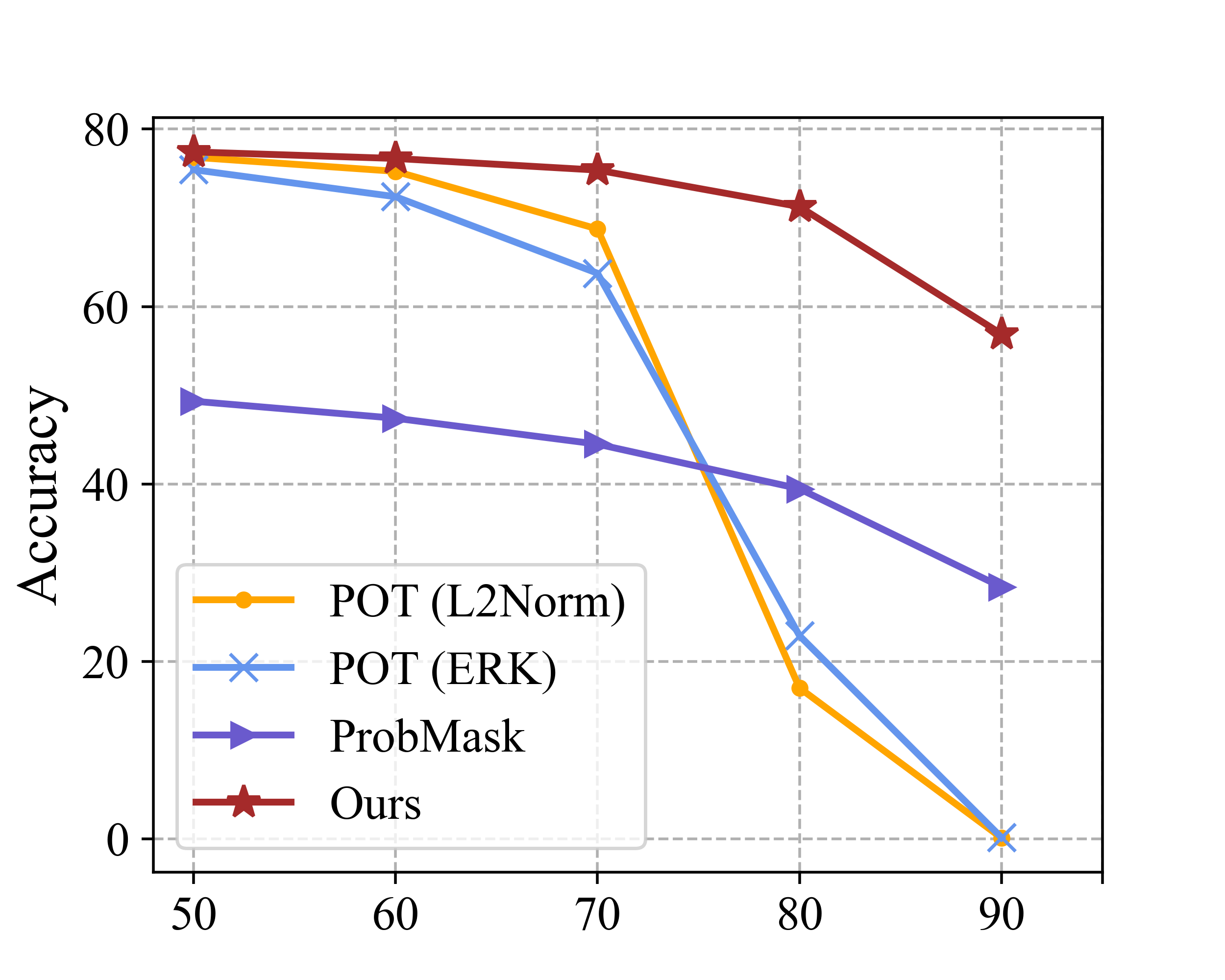}
	\end{center}
	\caption{Comparison with existing post-training sparsity methods on ImageNet ResNet-50. Our FCPTS enjoys significant accuracy improvement, especially for the extremely high sparsity rates (e.g., ~30\% boost under 80\% sparsity).}
	\label{fig:contribution}
\end{figure}
However, despite the benefit of inference performance for model sparsification, recovering the accuracy of sparse models is non-trivial. Most methods need to retrain the model for a long time with a large amount of data. It usually takes several hours and even several days when the training dataset is huge, bringing obstacles for large-scale applications in all walks of life. Recently, POT~\cite{lazarevich_post-training_2021} proposes to sparsify a neural network in a post-training way without training labels, significantly decreasing the cost of producing a sparse model. They reconstruct the sparse weight layer by layer to improve the accuracy. However,  to achieve comparable performance with the dense counterpart, its maximal overall sparsity rate can only reach 50\%. When the sparsity level increases, accuracy will crash quickly. This reveals that pushing the limit of post-training sparsity to a higher sparsity level is still challenging. 
To overcome this challenge, we first identified that the bottleneck of current PTS methods lies in sparsity rate allocation. 
The naive non-uniform sparsity allocation does not sufficiently utilize the sparsity sensitivity of each layer. Thus the network performance will degrade greatly as long as the sparse rate exceeds the highest tolerable level that the most sensitive layer can accept. 

To solve this problem, many traditional retraining-based approaches are proposed to generate a sparsity rate for each layer, which can be categorized into (1) empirical methods like ERK~\cite{rigl}, LAMP~\cite{lamp}, etc. (2) learning-based methods like STR~\cite{str}, LTP~\cite{ltp}, etc. Both types of methods encounter problems under the PTS setting. The empirical methods heavily depend on handicraft-designed prior knowledge and cannot promise an optimal solution. The learning-based methods either destroy the original weight distribution or require end-to-end training for convergence. Thereby, they can not realize the efficient and accurate sparsity allocation for the post-training case. What's more, many of them need to carefully adjust the hyperparameter and repeat multiple experiments to reach a target sparsity rate, further improving the costs of producing sparse models.

So, the problem remains: \textit{How to design an efficient and effective approach for post-training sparsity?} To address this issue, we propose the Fast and Controllable Post-training Sparsity (FCPTS) framework, which can learn the optimal sparsity allocation with a global constraint in minutes. Specifically, we set up a bridge between the pruning threshold and sparsity rate from the probability density perspective. It is non-trivial to build such a bridge because the sparsity rate calculation is non-differentiable, hindering the backpropagation for sparsity allocation learning. To this end, we use the Kernel Density Estimation (KDE) technique to build this bridge, which can be represented as a differentiable format, making the backpropagation possible. In this way, the sparsity rate of layers across the whole net can be directly optimized in a net-wise pipeline, which is more efficient than the layer-wise pipeline in the work \cite{lazarevich_post-training_2021}. 

Benefiting from the controllable net-wise optimization, we can obtain the desired sparse neural network in one pass. The whole reconstruction process can be completed within 30 minutes for typical DNNs (e.g., ResNet-18). What's more, equipped with the differentiable modeling of sparsity rate, we can learn the optimal sparsity allocation for each layer without harm to the original weight and thus further unleash the potential for higher sparsity level. With FCPTS, we can produce models with a 70\% global sparsity rate with accuracy on par with the dense counterparts. The total process just takes dozens of minutes. 

\begin{figure*}[tp!]
	\begin{center}
		\includegraphics[width=0.85\linewidth]{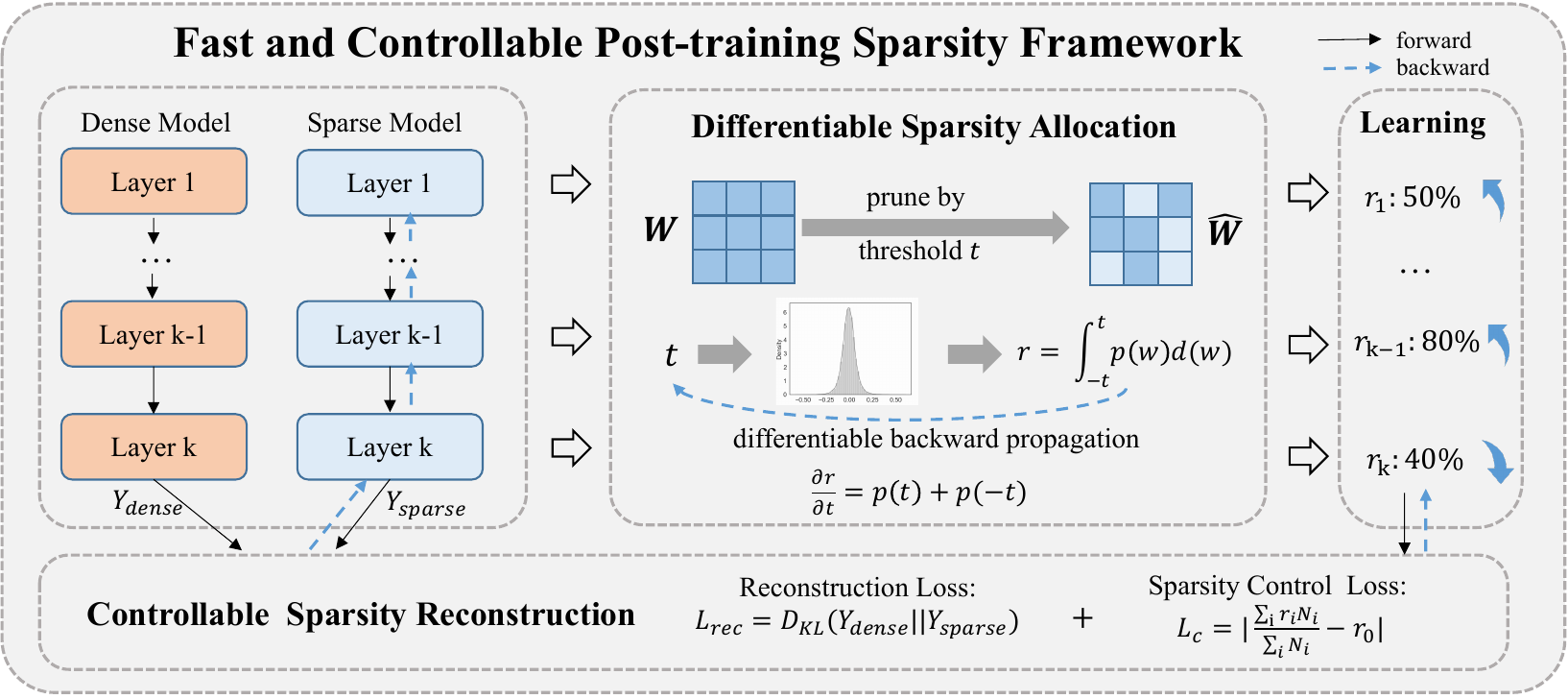}
	\end{center}
	\caption{An overview of our fast and controllable post-training sparsity (FCPTS) framework. The differentiable sparsity allocation transfers the learning of the sparsity rate to the threshold by a differentiable estimation, and the controllable sparsity reconstruction enables an optimized network with a specified global sparsity rate. Both components contribute to the final excellent performance.}
	\label{fig:overview}
\end{figure*}

Our FCPTS framework (Figure \ref{fig:overview}) owns the following advantages compared to state-of-the-art solutions:
\begin{enumerate}
    \item High efficiency. With FCPTS, the reconstructed sparse network can converge quickly with limited data by net-wise optimization. Thanks to the post-training paradigm and the net-wise optimization pipeline, our FCPTS is extremely efficient. To the best of our knowledge, the time for producing accurate sparse ImageNet models is for the first time reduced to minutes, compared to hours for the existing PTS method. 
    \item Controllable sparsity allocation. With the help of differentiable FCPTS, the final optimized sparse neural network can promise a specified sparsity rate, releasing the efforts of complex hyperparameter tuning.
    \item Accurate and optimal. With the optimized sparsity allocation, we can fully unleash the sparsity potential of each layer in the network and thus generate more accurate sparse models under the same global sparsity rate.
    \item Simple and general. FCPTS is easy to implement as a plugin and generalizes on various neural network architectures (ResNet, MobileNet, RegNet, ViT) and tasks (CIFAR-10/100, ImageNet, and PASCAL VOC).
\end{enumerate}

\section{Related Works}
Model sparsity dates back decades of years to \shortcite{thimm1995evaluating} which proved that pruning weights based on magnitude was a simple and powerful approach. Nowadays, there are two types of sparsity: structural and non-structural sparsity. In this paper, we focus on the latter one which has a more general format. As for the non-structural sparse models, current methods are mostly based on retraining. Only a few methods sparsify a trained model in the post-training scheme.

\emph{Retraining-based and Post-training Sparsity. }
Various retraining-based methods are proposed in the literature~\cite{strom1997sparse,han2015learning,molchanov2017variational,frankle2018the,renda2020comparing,narang2017exploring,guo2020model,guo2023multidimensional,Huang_2023_CVPR} to improve the accuracy. Although some of them may show promise, they often require extensive training or tuning of hyper-parameters, leading to inefficiency and slow convergence. Recently, a post-training sparsity method~\cite{lazarevich_post-training_2021} was proposed, achieving around 50\% sparsity rate without significant accuracy degradation. However, its accuracy decreases sharply at higher sparsity levels.

\emph{Uniform and Non-uniform Sparsity. }
The sparsity allocation methods can be categorized as uniform sparsity and non-uniform sparsity. Uniform sparsity methods~\cite{pmlr-v97-mostafa19a, Lin2020Dynamic, gale2019state} allocate the same sparsity rate for all the layers, while non-uniform ones assign lower sparsity rates to sensitive layers. Although empirical methods~\cite{rigl,frankle2018the,renda2020comparing} have been used to achieve non-uniform sparsity, they cannot guarantee optimal results as the pruning threshold and sparsity allocation are not jointly learned. Thus there are also methods that aim to learn sparsity by optimizing the mask or pruning threshold, but they all require a long training time and do not work well in post-training settings.

\section{Fast and Controllable \\Post-training Sparsity Framework}
In this section, we introduce our Fast and Controllable Post-training Sparsity (FCPTS) framework. We first define the controllable sparsity reconstruction objective, and then design a differentiable sparsity allocation method without disturbing the trained weights. Based on these, the optimal sparsity allocation can be easily learned.
\subsection{Preliminary}

\emph{Mask Generation. } Usually, magnitude-based methods have been proven to be simple and effective metrics for mask selection, which can be formulated as follows:
\begin{equation}
\label{eq:mask_generation}
    \mM = 0.5 * \text{sgn}(|\mW| - t) + 0.5,
\end{equation}
where $\text{sgn}$ is the sign function which returns $+1$ for positive values and otherwise returns $-1$. $t$ is a threshold for pruning weights. Then the sparse model's output can be derived by:
\begin{equation}
\label{eq:nn_function}
    \mY_{sparse} = f(\mX, \mM \odot \mW).
\end{equation}
where $f$ denotes the neural network. The mask generation method in Equation \ref{eq:mask_generation} implies the fact that weights with larger magnitudes have higher importance for accuracy. This assertion indeed holds for weights in an individual layer.

However, for weights across layers in the whole network, we can not compare them directly due to the huge difference in weight norm. In other words, different layers also have different contributions to the final accuracy. To this end, finding the most suitable sparsity rate for each layer becomes important, especially for the post-training scenario.

\subsection{Controllable Sparsity Reconstruction}
In this part, we present our formulation for controllable sparsity reconstruction, which defines the optimization objective of our FCPTS framework. It incorporates a control loss to achieve a target sparsity rate and employs reconstruction supervision to minimize the difference between the sparse output and dense counterpart.

First, we will introduce the control loss. As mentioned earlier, existing methods mainly use regularization to adjust the weight magnitude, then indirectly influence the sparsity rate. This results in a complicated process to achieve the desired sparsity rate and prevents us from efficiently obtaining a sparse neural network in a limited time. To address this problem, we construct a control loss as defined as:
\begin{equation}
    \label{eq:control_loss}
    L_{c} = |\frac{\sum_{i} r_i N_i}{\sum_{i} N_i} - r_0|,
\end{equation}
where the $r_i$ denotes the sparsity rate and $N_i$ is element number of weights of the $i_{th}$ layer. $r_0$ is the global sparsity rate target and $L_{c}$ is the loss for controlling the global sparsity rate. With this objective $L_c$, we can reach the target sparsity rate $r_0$ without complex hyperparameter tuning.

Furthermore, accompanied by the control loss, we employ a weight reconstruction technique, commonly used in post-training quantization~\cite{2022-iclr-qdrop}, to reduce the difference between sparse and dense output.
\begin{equation}
    \label{eq:rec_loss}
    L_{rec} = \KL(\mY_{dense} \| \mY_{sparse}), \\
\end{equation}
$\KL(\cdot)$ represents the Kullback–Leibler divergence function. Under the supervision of the reconstruction loss $L_{rec}$, the weight optimization will be guided in a direction that contributes to a sparse output closely resembling the dense output and naturally a higher accuracy. What's more, the overall reconstruction is based on the well-trained dense weights, and thus the process is fast and easy to converge.

Finally, we combine $L_c$ and $L_{rec}$ to determine the appropriate sparsity rate for each layer while also making minor adjustments to the weights to accommodate the sparsity. The overall optimization objective can be defined as:
\begin{equation}
    L = L_{rec} + L_{c},
\end{equation}
Guided by the whole objective, we can find the reconstructed weight $\mW$ and sparsity rate $r_i$ of each layer that best suits the target sparsity rate $r_0$. 
\begin{equation}
    \argmin_{\mW, r_i} L
\end{equation}

\subsection{Differentiable Sparsity Allocation}
After the above optimization objective is defined, we need to make the optimization process possible. To realize this, an ingenious bridge function is designed to calculate the sparsity rate $r$ from the magnitude threshold $t$. The derivative of this bridge function can be calculated by kernel density estimation. Then the differentiable optimization of sparsity allocation can be achieved by transferring the optimization to $t$. With the optimal $t$ of each layer being learned, the sparsity rate $r$ of each layer can be obtained.

Given a specific layer $l$, the derivative of $L$ with respect to $r_l$ is needed to fulfill the differentiable sparsity learning.
\begin{equation}
    \label{eq:loss_derivative}
    \frac{\partial L}{\partial r_l} = \frac{\partial L_{rec}}{\partial r_l} + \frac{\partial L_c}{\partial r_l}.
\end{equation}
For the latter part of Equation \ref{eq:loss_derivative}, it can be easily obtained according to Equation \ref{eq:control_loss}.
\begin{equation}
    \label{eq:control_derivative}
    \frac{\partial L_{c}}{\partial r_l} = \begin{cases}
    \frac{N_l}{\sum_{i} N_i}, &\text{if} \frac{\sum_{i} r_i N_i}{\sum_{i} N_i} > r_0,\\
    -\frac{N_l}{\sum_{i} N_i}, &\text{otherwise}.
    \end{cases}
\end{equation}
Then we mainly focus on the former part. 
\begin{align}
\label{eq:rec_derivative}
    \frac{\partial L_{rec}}{\partial r_l} &= \frac{\partial L_{rec}}{\partial \mM_l} \cdot \frac{\partial \mM_l}{\partial t_l} \cdot \frac{\partial t_l}{\partial r_l}
\end{align}
Combined with Equation \ref{eq:mask_generation}, Equation \ref{eq:nn_function}, and Equation \ref{eq:rec_loss}, the first two items $\frac{\partial L_{rec}}{\partial \mM_l}$ and $\frac{\partial \mM_l}{\partial t_l}$ are easy to calculate. Thereby the core problem is to calculate $\frac{\partial t_l}{\partial r_l}$. If we can find a differentiable function $g$ that lets $ t_l = g(r_l)$. Then the computation of Equation \ref{eq:rec_derivative} will be feasible.

\textbf{\emph{Bridge function $g^{-1}$ from $t_l$ to $r_l$: }} Unfortunately, a differentiable $g$ is hard to construct. But the inverse function $g^{-1}$ can be formalized in a format with differentiable estimation. Then we can transfer the learning of sparsity rate $r$ to threshold $t$. Given a weight distribution $\mW_l$ of layer $l$ and its probability density function (PDF) $p$, $w$ is sampled from the distribution and then the sparsity rate $r_l$ can be got by:
\begin{equation}
\label{eq:bridge_function}
    r_l = \int_{-t_l}^{t_l}p(w)d(w) = g^{-1}(t_l).
\end{equation}
\begin{figure}[htbp]
	\begin{center}
	\includegraphics[width=0.9\linewidth]{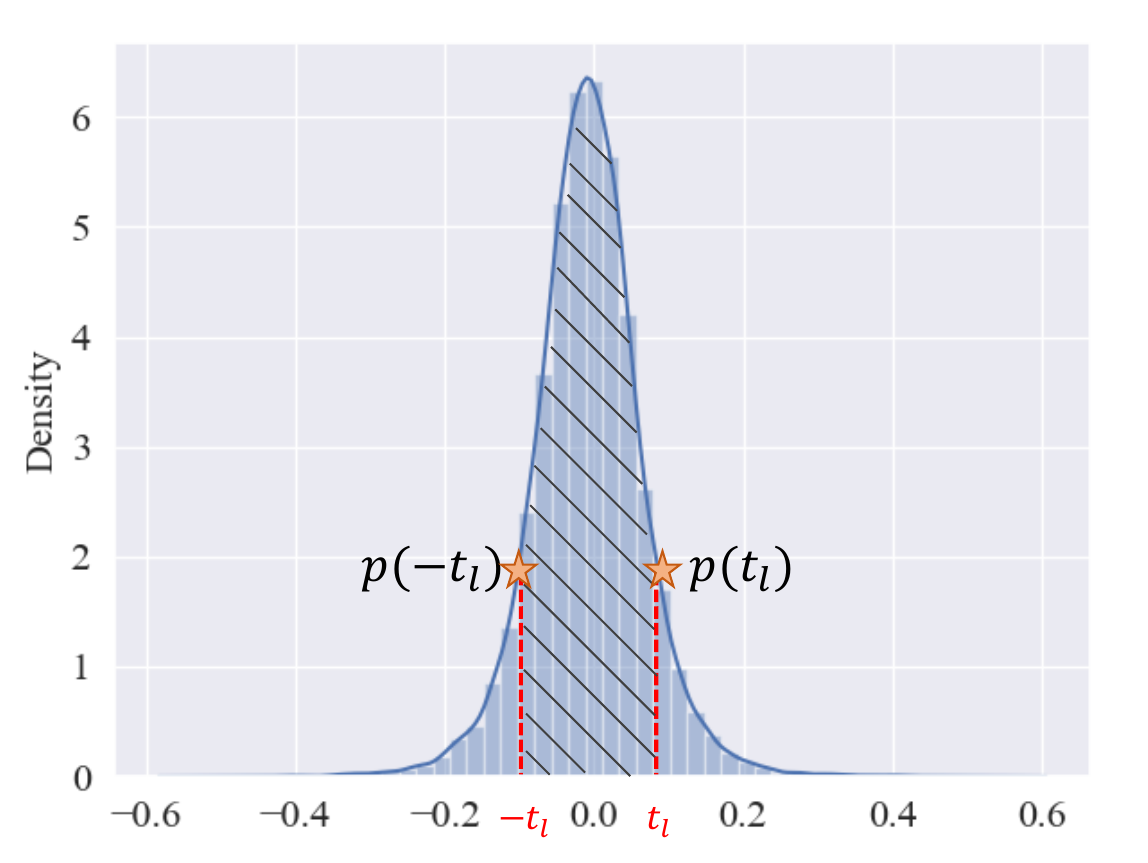}
	\end{center}
	\caption{Explanation of bridge function $g^{-1}$ from threshold $t_l$ to sparsity $r_l$ (Equation \ref{eq:bridge_function}). The area of the shading region equals the sparsity rate $r_l$. The derivative of $r$ with respect to $t$ can be represented as $p(-t_l) + p(t_l)$.}
	\label{fig:bridge_function}
\end{figure} 

Thus the derivative of $t_l$ with respect to $r_l$ can be written as:
\begin{equation}
\label{eq:bridge_derivative}
    \frac{\partial r_l}{\partial t_l} = p(t_l) + p(-t_l).
\end{equation}

\textbf{\emph{Differentiable Estimation:}} From Equation \ref{eq:bridge_derivative}, the key step to calculate the derivative is modeling the probability distribution $p$. To this end, we utilize the kernel density estimation (KDE) method to estimate the PDF $p$.
\begin{equation}
\label{eq:kde}
    p(w) = \frac{1}{n} \sum_{i}^{n}K_h(w-w_i) = \frac{1}{nh}\sum_{i}^nK(\frac{w-w_i}{h}).
\end{equation}
where $K$ is a non-negative kernel function. $n$ is the number of sampled points. $h>0$ is a smoothing parameter called the bandwidth. A kernel with subscript $h$ is called the scaled kernel and defined as $K_h(w) = \frac{1}{h} K(\frac{w}{h})$. Intuitively we want to choose $h$ as small as the data will allow. However, there is always a trade-off between the bias of the estimator and its variance. Here we adopt a commonly used setting: $n=100, h=0.5$ and $K(x) = \Phi(x)$, where $\Phi$ is the standard normal density function. With this KDE technique, the bridge function is proved to be differentiable.

\textbf{\emph{Transfer the learning of $r$ to $t$: }} Equipped with the above bridge function and differentiable estimation, we can optimize the sparsity rate $r$ via the proxy variable $t$:
\begin{equation}
    \frac{\partial L_{c}}{\partial t_l} = \frac{\partial L_{c}}{\partial r_l} \cdot \frac{\partial r_l}{\partial t_l}.
\end{equation}
According to Equation \ref{eq:control_derivative}, Equation \ref{eq:bridge_derivative} and Equation \ref{eq:kde}, the derivative of $L_c$ with respect to $t_l$ can be obtained.
As for the reconstruction loss, the derivative of $L_{rec}$ with respect to $t_l$ can also be directly calculated:
\begin{equation}
    \frac{\partial L_{rec}}{\partial t_l} = \frac{\partial L_{rec}}{\partial \mM_l} \cdot \frac{\partial \mM_l}{\partial t_l}.
\end{equation}

Finally, with the learned threshold $t$ and bridge function Equation \ref{eq:bridge_function}, we can calculate the exact $r$, contributing to the flexible and controllable learning of sparsity allocation.

\subsection{Discussion}
\label{Discussion}
Benefiting from the controllable and differentiable sparsity allocation, our FCPTS performs well for the post-training scenario and can reconstruct a sparse neural network with high accuracy as quickly as possible.

\textit{\textbf{Superiority over existing retraining-based non-uniform methods. }} STR~\shortcite{str} utilizes $\hat{\mW} = \textit{relu}(\mW - t)$ to learn the threshold $t$. However, for weights $>t$, the original weight value will be damaged (subtracted by $t$), hindering the fast accuracy recovery. In contrast, our method will fully utilize the dense weights without damaging their original distribution, contributing to an effective post-training reconstruction. Also STR can not constrain the global sparsity rate into the target value we want. Without careful regularization tuning, the sparsity rate could quickly spiral out of control. ProbMask~\cite{zhou2021effective} optimizes the mask with Gumbel Softmax trick. The large optimization space and the stochastic optimization method require extensive steps and huge GPU memory to stably converge. For post-training sparsity, ProbMask is unstable and usually fails to converge to a good solution in a limited time. Table \ref{tab:learning_fail} illustrates experimental evidence about the poor performance of existing methods under the PTS setting. The two retraining-based methods STR and ProbMask even underperform the naive post-training method POT~\cite{lazarevich_post-training_2021}.

\begin{table}[]
\begin{center}
\small
\begin{tabular}{cccc}
\hline
  & \multicolumn{3}{c}{Sparsity Rate(\%)} \\ \cline{2-4} \multirow{-2}{*}{Method} & 50 & 60 & 70\\ \hline
  POT & 70.1/90.12 & 68.40/89.10 & 63.72/86.51\\ 
  STR & \XSolidBrush & \XSolidBrush & \XSolidBrush \\
  ProbMask & 51.97/79.65 & 49.64/78.19 & 46.99/76.29 \\ \hline
\end{tabular}
\end{center}
\caption{Top1/Top5 accuracy of ResNet-18 on ImageNet with different sparsity rates. The traditional learning-based non-uniform sparsity methods fail to work well for the post-training setting, either crashing or converging unstably.}
\label{tab:learning_fail}
\end{table}

\textbf{Efficiency. } FCPTS enjoys high efficiency since it reaches the target sparsity rate with only one pass of net-wise reconstruction, instead of layer-by-layer progressive optimization in POT. Usually, it can generate sparse neural networks in dozens of minutes using just one NVIDIA RTX 3090 GPU. The overall pipeline is summarized in Algorithm \ref{alg}.

\begin{algorithm}[t]
    \caption{FCPTS Framework.}
    \label{alg}
    \KwIn{Calibration dataset $\mathcal{D}$, target sparsity rate $r_0$, a well-trained dense network with weight $\mW$.} 
    \For{$d \in \mathcal{D}$}{
        \{1. Forward propagation:\}\\
        $\mM = 0.5 * \textit{sgn}(|\mW|-t)+0.5$,\\
        $\hat{\mW}=\mM\odot\mW$, \\
        $\mY_{dense} = f(\mX, \mW), \mY_{sparse} = f(\mX, \hat{\mW})$, \\
        Calculate the $L_{rec}$ by Equation \ref{eq:rec_loss}; \\
        Obtain the sparsity rate $r$ by $t$ using Equation \ref{eq:bridge_function}, \\
        Calculate the $L_c$ by Equation \ref{eq:control_loss}; \\
        Get the overall $L = L_{rec} + L_c$.\\
        
	\{2. Backward propagation:\}\\
        Calculate $p(w)$ by kernel density estimation in Equation \ref{eq:kde}, \\
        Obtain the derivative of $r_l$ with respect to $t_l$ by Equation \ref{eq:bridge_derivative}, \\
	Compute the gradient of $t$ by Equation \ref{eq:control_derivative} and Equation \ref{eq:rec_derivative}, \\
        Adjust the weight and sparsity rate by gradient descent \;
    }
    \textbf{return} sparse networks with target sparsity rate $r_0$ \;
\end{algorithm}

\section{Experiment}
In this section, we conduct a series of experiments to evaluate the effect of FCPTS. We first conduct extensive experiments on image classification and object detection to compare with existing methods. Then in-depth analyses are given to reveal the internal superiority of our method.  
\begin{table*}[t!] 
\small
\begin{center}
\begin{tabular}{cccccccc}
\hline
\multirow{2}{*}{Model}     & \multirow{2}{*}{Method} & \multicolumn{3}{c}{CIFAR-10}            & \multicolumn{3}{c}{CIFAR-100}           \\ \cline{3-8} 
                           &                         & 70          & 80          & 90          & 70          & 80          & 90          \\ \hline
\multirow{3}{*}{ResNet-32} & POT (L2Norm)            & 91.06/99.64 & 84.80/98.85 & 20.09/71.73 & 59.35/86.47 & 28.98/63.09  & 3.01/11.55   \\
                           & POT (ERK)               & 90.95/99.62 & 84.91/98.59 & 21.87/74.36 & 56.21/84.71 & 29.50/62.64  & 4.66/14.30   \\
                           & Ours                    & \textbf{92.91/99.77} & \textbf{92.36/99.73} & \textbf{90.35/99.65} & \textbf{69.14/91.07} & \textbf{68.28/90.87} & \textbf{63.35/89.38} \\ \hline
\multirow{3}{*}{ResNet-56} & POT (L2Norm)            & 93.01/99.74 & 87.18/98.83 & 28.21/72.24 & 64.63/88.90 & 36.19/69.07  & 4.41/13.00   \\
                           & POT (ERK)               & 92.84/99.70 & 87.03/98.47 & 37.70/70.33 & 62.65/88.16 & 36.96/69.44  & 5.88/17.37   \\
                           & Ours                    & \textbf{94.07/99.83}          & \textbf{93.67/99.78}           & \textbf{92.02/99.78}           & \textbf{72.16/92.02} & \textbf{71.09/91.90} & \textbf{68.07/90.51} \\ \hline
\end{tabular}
\end{center}
\caption{Comparison of the Top1/Top5 accuracy (\%) on ResNet-32/56 under different sparsity rates on dataset CIFAR-10/100. The accuracies (\%) of dense ResNet-32 on dataset CIFAR-10/100 are 93.53/99.77 and 70.16/90.89, respectively. The accuracies (\%) of dense ResNet-56 on dataset CIFAR-10/100 are 94.37/99.83 and 72.63/91.94, respectively.}
\label{CIFAR}
\end{table*}

\begin{table*}[tp!]
\small
\begin{center}
\begin{tabular}{ccccccc}
\hline
\multirow{2}{*}{Model}                                                              & \multirow{2}{*}{Method} & \multicolumn{5}{c}{Sparsity Rate (\%)}                                                                                                  \\ \cline{3-7} 
                                                                                    &                         & 50                   & 60                   & 70                   & 80                   & 90                                      \\ \hline
\multirow{4}{*}{\begin{tabular}[c]{@{}c@{}}ResNet-18\\ 70.88/90.45\end{tabular}}    & POT (L2Norm)            & 70.06/89.12 & 68.40/89.10          & 63.72/86.51          & 44.94/71.51          & 6.03/15.70                      \\
                                                                                    & POT (ERK)               & 69.66/89.40          & 68.24/88.62          & 64.28/86.40          & 50.78/75.59          & 5.66/15.81                      \\
                                                                                    & ProbMask                & 51.97/79.65          & 49.64/78.19          & 46.99/76.29          & 42.58/72.58          & 32.88/64.33                   \\
                                                                                    & Ours                    & \textbf{70.07/89.61}          & \textbf{69.58/89.37} & \textbf{68.50/88.89} & \textbf{65.83/87.69} & \textbf{57.40/83.37} \\ \hline
\multirow{4}{*}{\begin{tabular}[c]{@{}c@{}}ResNet-50\\ 77.89/93.762\end{tabular}}   & POT (L2Norm)            & 76.83/93.81 & 75.24/93.22          & 68.74/89.44          & 17.03/30.31          & 0.13/0.42              \\
                                                                                    & POT (ERK)               & 75.43/92.51          & 72.38/90.89          & 63.76/84.74          & 22.92/42.13          & 0.17/1.25                 \\
                                                                                    & ProbMask                & 49.34/78.42          & 47.42/76.88          & 44.50/74.95          & 39.47/70.97          & 28.40/60.22       \\
                                                                                    & Ours                    & \textbf{77.43/93.55}          & \textbf{76.69/93.38} & \textbf{75.38/92.87} & \textbf{71.26/91.23} & \textbf{56.91/84.86}\\ \hline
\multirow{4}{*}{\begin{tabular}[c]{@{}c@{}}RegNetX-200M\\ 68.41/89.11\end{tabular}} & POT (L2Norm)            & 64.69/87.08          & 59.98/84.40          & 48.36/75.86          & 24.48/49.90          & 2.02/7.42                 \\
                                                                                    & POT (ERK)               & 64.54/86.44          & 60.71/84.40          & 52.97/79.46          & 31.56/60.02          & 1.36/4.83                   \\
                                                                                    & ProbMask                & 50.23/77.66          & 47.67/75.97          & 44.61/73.71          & 39.43/69.78          & 29.87/60.26          \\
                                                                                    & Ours                    & \textbf{66.57/87.30} & \textbf{65.30/86.73} & \textbf{62.80/85.32} & \textbf{57.44/82.37} & \textbf{40.90/71.47}     \\ \hline
\multirow{4}{*}{\begin{tabular}[c]{@{}c@{}}RegNetX-400M\\ 71.84/90.55\end{tabular}} & POT (L2Norm)            & 67.54/88.62          & 65.68/87.73          & 56.63/81.66         & 28.55/54.41          & 2.32/7.87                  \\
                                                                                    & POT (ERK)               & 69.97/89.59          & 67.07/88.25          & 60.40/84.22          & 37.10/65.71          & 1.23/3.53                \\
                                                                                    & ProbMask                & 48.84/77.17          & 46.25/75.54          & 43.20/73.15          & 38.59/69.14          & 30.30/60.96           \\
                                                                                    & Ours                    & \textbf{70.85/90.17} & \textbf{70.09/89.73} & \textbf{68.27/89.13} & \textbf{64.23/87.15} & \textbf{50.00/78.97}\\ \hline
\multirow{4}{*}{\begin{tabular}[c]{@{}c@{}}MobileNetV2\\ 72.85/91.61\end{tabular}}  & POT (L2Norm)            & 65.51/87.31          & 57.95/81.35          & 21.60/44.53          & 0.22/1.47            & 0.10/0.55           \\
                                                                                    & POT (ERK)               & 69.80/89.66 & 64.98/87.14          & 49.14/76.25          & 9.75/25.30           & 0.21/0.77            \\
                                                                                    & ProbMask                & 30.24/63.68          & 25.15/57.52          & 18.92/48.66          & 11.76/35.48          & 4.41/16.75           \\
                                                                                    & Ours                    & \textbf{70.52/90.07}          & \textbf{68.10/89.16} & \textbf{61.18/86.04} & \textbf{40.20/74.28} & \textbf{13.99/39.75}  \\ \hline
\end{tabular}
\end{center}
\caption{Comparison of the Top1/Top5 accuracy (\%) on various CNNs under different sparsity rates on ImageNet dataset. The accuracy (\%) of the dense model is listed under the architecture of the model.}
\label{ImageNet-1K}
\end{table*}

\begin{table}[]
\setlength\tabcolsep{1pt}
\small
\begin{center}
\begin{tabular}{cclll}
\hline
\multirow{2}{*}{Model}                                                           & \multirow{2}{*}{Method} & \multicolumn{3}{c}{Sparsity Rate (\%)}                                   \\ \cline{3-5} 
                                                                                 &                         & \multicolumn{1}{c}{50} & \multicolumn{1}{c}{60} & \multicolumn{1}{c}{70} \\ \hline
\multirow{3}{*}{\begin{tabular}[c]{@{}c@{}}ViT base\\ 75.68/92.94\end{tabular}}  & POT(L2Norm)             & 62.08/83.92            & 53.63/77.08            & 29.96/53.16            \\
                                                                                 & POT(ERK)                & 68.04/89.91            & 61.67/71.48            & 30.50/55.30            \\
                                                                                 & Ours                    & \textbf{74.90/92.99}            & \textbf{72.09/91.47}            & \textbf{65.24/87.74}            \\ \hline
\multirow{3}{*}{\begin{tabular}[c]{@{}c@{}}ViT large\\ 79.29/94.78\end{tabular}} & POT(L2Norm)             & 77.11/93.91            & 75.33/93.06            & 70.42/90.78            \\
                                                                                 & POT(ERK)                & 77.46/94.22            & 75.64/93.33            & 71.02/91.23            \\
                                                                                 & Ours                    & \textbf{78.13/94.41 }           & \textbf{76.71/93.73}            & \textbf{73.01/92.07}            \\ \hline
\end{tabular}
\end{center}
\caption{Comparison of the Top1/Top5 accuracy (\%) on ViT models under different sparsity rates on ImageNet dataset.}
\label{ViT}
\end{table}

\subsection{Results on Various Tasks}
We conduct comprehensive validations on four datasets covering image classification (CIFAR-10/100~\cite{cifar}, ImageNet~\cite{imagenet}) and object detection (PASCAL VOC~\cite{journals/ijcv/EveringhamGWWZ10}). The CIFAR-10 dataset consists of 50K training images and 10K testing images of size 32$\times$32 with 10 classes. ImageNet ILSVRC12 contains about 1.2 million training images and 50K testing images with 1,000 classes. The PASCAL VOC is a widely used object detection dataset containing 20 object categories. Each image in this dataset has bounding box annotations and object class annotations.

\noindent\textbf{Baseline Setting.} We choose the only PTS method POT~\cite{lazarevich_post-training_2021} as our baseline. The calibration dataset contains 10k images. Also, we reproduced some techniques originally designed for retraining, i.e., the heuristic non-uniform sparsity method ERK in POT, and ProbMask under the PTS setting.

\noindent\textbf{Network Structures.} We employ the widely-used network structures including ResNet-32/56 for CIFAR-10/CIFAR-100, and ResNet-18/50~\cite{resnet}, MobileNetV2~\cite{mbv2}, RegNet-200M/400M~\cite{regnet}, Vit-Base/Large~\cite{vit} for ImageNet, and MobileNetV1 SSD, MobileNetV2 SSD-lite~\cite{liu2016ssd} for PASCAL VOC.

\subsubsection{CIFAR-10/CIFAR-100}

Table \ref{CIFAR} clearly shows that our method outperforms the baselines. As the sparsity increases, the superiority becomes even more significant. The baseline method degrades rapidly at an 80\% sparsity and collapses completely at 90\% or earlier. In contrast, our FCPTS remains stable at a very high accuracy even at a sparsity of 90\%. This is particularly evident in the CIFAR-100 dataset.


\subsubsection{ImageNet}

The accuracy results for various sparsity rates on ImageNet are presented in Table \ref{ImageNet-1K}. Our method is shown to outperform the baseline with a significant margin across all models, particularly at sparsity rates exceeding 70\%. Notably, at a sparsity rate of 70\%, our method achieves accuracy levels almost similar to its dense counterpart on ResNet-18 and ResNet-50 (with a mere 2\% degradation), while existing methods experience a significant accuracy loss of near 10\%. For mobile-friendly architectures like RegNet and MobileNet, existing methods encounter unacceptable accuracy decreases under 60\% sparsity, whereas our method can improve by 3\%-8\%. These results provide detailed evidence of the advantages of our proposed method. Furthermore, we tested the effect on ViT models in Table \ref{ViT}. The consistent improvement proves that our FCPTS also generalizes for attention-based architectures.


\begin{table}[tp!]
\small
\begin{center}
\begin{tabular}{ccc}
\hline
Model & Method & mAP (\%) \\ \hline
 & POT (L2Norm) & {\color[HTML]{1F2329} 48.5 } \\
 & POT (ERK) & {\color[HTML]{1F2329} 54.6} \\
\multirow{-3}{*}{\begin{tabular}[c]{@{}c@{}}MobileNetV1 SSD\\ 67.7\end{tabular}} & Ours & \textbf{65.1} \\ \hline
 & POT (L2Norm) & 16.4 \\
 & POT (ERK) & 0.3 \\
\multirow{-3}{*}{\begin{tabular}[c]{@{}c@{}}MobileNetV2 SSD-Lite\\ 68.6\end{tabular}} & Ours & \textbf{59.1} \\ \hline
\end{tabular}
\end{center}
\caption{Comparison on PASCAL VOC at 90\% sparsity.}
\label{VOC}
\end{table}
\subsubsection{PASCAL VOC}

We report the detailed accuracy from two variants of SSD at the sparsity rate 90\% in Table \ref{VOC}. Compared with the POT using L2-normalization magnitude and ERK sparsity rate allocation algorithm, our FCPTS can get a significant improvement under this extreme conditions of 90\% sparsity rate. For the sparse MobileNetV1 SSD, the mAP of our FCPTS only drops a little. For the compressed MobileNetV2 SSD-Lite model, our FCPTS can get 59.1\% mAP while the performances of other methods crash.

\begin{figure}[tp!]
	\begin{center}
		\includegraphics[width=1\linewidth]{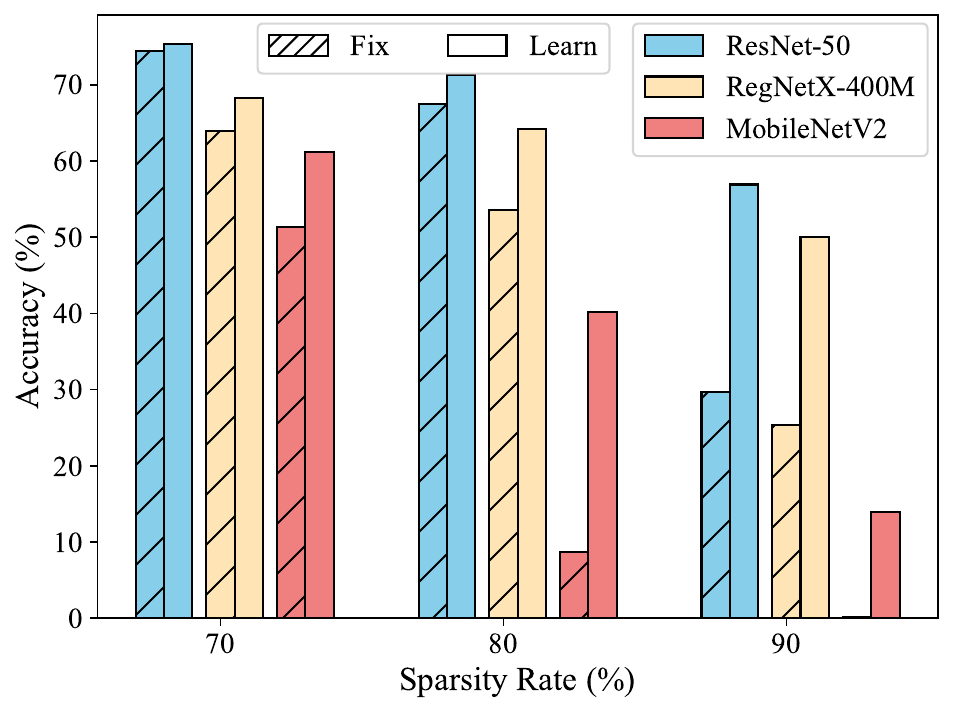}
	\end{center}
        \caption{The effect of learnable sparsity. For each neural network, the left dash bars are results with fixed sparsity rate and the right bars are results with learned sparsity rate.}
	\label{fig:ablation_study}
\end{figure} 
\subsection{Effect of Learnable Sparsity}
\label{Effect of Learnable Sparsity}
In this section, we study how the learnable sparsity contributes to the final performance. We just initialize the sparsity allocation of models with ERK and fix the sparsity rates of all layers, then reconstruct with or without learning the sparsity rate. We conducted this experiment on three models, ResNet-50, RegNetX-400M, and MobileNetV2. Figure \ref{fig:ablation_study} shows that models with learnable sparsity can get higher accuracy compared with models without learnable sparsity. This gap in accuracy is more pronounced at high sparsity rates, e.g., about 10\% improvement for RegNetX-400M under the sparsity rate of 80\%. This experiment strongly demonstrates that the learnable sparsity with a global constraint can dynamically optimize the sparsity allocation to make it more reasonable and rescue it from some unexpected sparsity allocation traps.


\subsection{Sparsity Allocation Analyses}

\begin{figure}[t!]
	\begin{center}
		\includegraphics[width=1\linewidth]{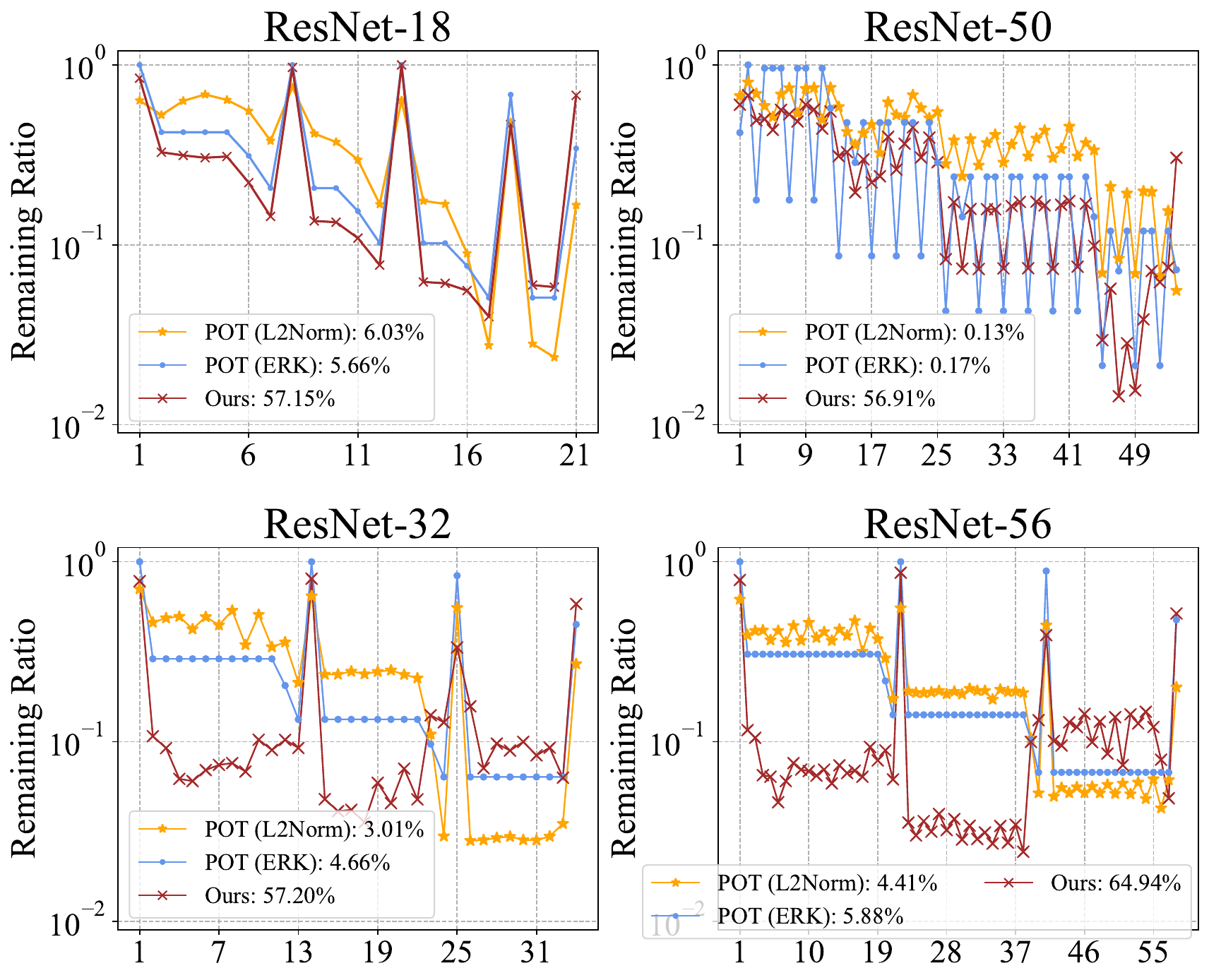}
	\end{center}
	\caption{Visualization of the optimized sparsity allocation at a sparsity rate of 90\%. ResNet-18 and ResNet-50 are on ImageNet, ResNet-32 and ResNet-56 are on CIFAR-100. }
	\label{fig:sparsity_allocation}
\end{figure} 
In Figure \ref{fig:sparsity_allocation}, we prove the effectiveness of FCPTS which can lead to more reasonable sparsity allocation. The optimized sparsity allocation for 4 models reconstructed on the ImageNet and CIFAR-100 datasets is presented. It can be found that our method effectively learns to allocate lower sparsity for the latter layers and increase sparsity for some other layers to reach the global sparsity rate target. Additionally, the first layer exhibits a relatively high sparsity rate, in alignment with experience in previous works. Without the learned allocation, the other two methods tend to exceed the sparsity tolerance and suffer a huge accuracy crash.
\begin{table}[htbp]
\begin{center}
\small
\begin{tabular}{cccc}
\hline
Method & Dataset & Model & Time (min) \\ \hline
  & CIFAR-100 & ResNet-32 & 31 \\ 
 \multirow{-2}{*}{POT} & ImageNet & ResNet-18 & 100 \\ \hline
 & CIFAR-100 & ResNet-32 & 9 \\ 
 \multirow{-2}{*}{Ours} & ImageNet & ResNet-18 & 29 \\
 \hline
\end{tabular}
\end{center}
\caption{Efficiency for generating a sparse neural network.}
\label{tab:reconstruction_efficiency}
\end{table}
\newcommand{\tabincell}[2]{\begin{tabular}{@{}#1@{}}#2\end{tabular}}
\begin{table}[htbp]
\small
\begin{center}
\begin{tabular}{ccccc}
\hline
Model & \tabincell{c}{Sparsity\\ rate (\%)}  &\tabincell{c}{Latency\\ (ms)} & \tabincell{c}{Speed\\up} & \tabincell{c}{Memory\\(MB)} \\ \hline
 & dense &  18.194 & - & 11.632 \\
 & 50 & 10.796 & 1.695 & 7.640 \\
 & 60 & 9.287 & 1.959 & 6.530 \\
\multirow{-3}{*}{ResNet-18} & 70 & 7.599 & 2.394 & 5.245 \\ \hline
 & dense & 33.936 & - & 6.207 \\
 & 50 & 28.387 & 1.195 & 6.488 \\
 & 60 & 28.157 & 1.205 & 5.686 \\
\multirow{-3}{*}{\tabincell{c}{MobileNetV2}} & 70 & 26.863 & 1.263 & 5.235 \\ \hline
\end{tabular}
\end{center}
\caption{Inference performance of sparse ResNet-18 and MobileNetV2 on CV22, a hardware of autonomous driving.}
\label{tab:inference_efficiency}
\end{table}


\subsection{Efficiency}
To present the high efficiency of our method, we collect the reconstruction time cost to obtain a sparse neural network and evaluate the inference speed on the real hardware.

\noindent\textbf{Reconstruction Efficiency. }
Since FCPTS obeys the original weight distribution of the dense model, it can fully take advantage of the historical training efforts. Besides, it directly adopts net-wise optimization on a reduced optimization space (i.e., threshold) without any randomness and without complex layer-wise calculation for MSE. So it can converge quickly and stably. More importantly, the controllable sparsity rate enables us to reach the target without repeated hyper-parameter tunings. From Table \ref{tab:reconstruction_efficiency}, we can see that our FCPTS for the first time pushes the time for generating ImageNet sparse models into minutes while the existing state-of-the-art method POT requires over one and a half hours. Our framework enjoys a 3 times speedup.

\noindent\textbf{Inference Efficiency. } We also test the inference performance of the models sparsified by FSPTS on Ambarella CV22, an autonomous driving chip supporting the acceleration of unstructured sparsity. As seen in Table \ref{tab:inference_efficiency}, benefiting from the sparsity, both inference latency and memory occupation are significantly reduced. This gain is especially remarkable for the ResNet-18, which achieves nearly 2.4x speedup and over 50\% memory saving at 70\% sparsity.

\section{Conclusion}
In this paper, we propose a fast and controllable post-training sparsity (FCPTS) framework that pushes the limit of PTS accuracy to a new level. It utilizes a differentiable estimation to enable a learnable and controllable sparsity rate. Benefiting from the optimal sparsity allocation, our FCPTS achieves state-of-the-art results on 4 different datasets covering classification and object detection tasks.

\section*{Acknowledgments}
This work was supported in part by the National Key Research and Development Plan of China (2022ZD0116405), the National Natural Science Foundation of China (No. 62206010, No.62022009, No. 62306025), and the State Key Laboratory of Software Development Environment (SKLSDE-2022ZX-23).

\bibliography{aaai24}

\end{document}